# Pixel Normalization from Numeric Data as Input to Neural Networks
## For Machine Learning and Image Processing

Parth Sane[1] and Ravindra Agrawal[2]
Department of Computer Engineering SIES GST, University of Mumbai Mumbai, India
Email: [1]parthsane@icloud.com, [2]agarwal.ravindra@siesgst.ac.in

*Abstract*—Text to image transformation for input to neural networks requires intermediate steps. This paper attempts to present a new approach to pixel normalization so as to convert textual data into image, suitable as input for neural networks. This method can be further improved by its Graphics Processing Unit (GPU) implementation to provide significant speedup in computational time.

## I. INTRODUCTION

Text to image transformation is nothing new as it is known that images can be represented with matrices having numeric elements. But, where our method differs in implementation is representing any numeric data into images. This is achieved by performing statistical normalization on entire dataset to be fed into destination process. Neural networks use databases as data source like Imagenet [1] to use images for classifying images into target classes. This inspired us to think of the churn problem visually. Churn rate (sometimes called attrition rate), in its broadest sense, is a measure of the number of individuals or items moving out of a collective group over a specific period. It is one of two primary factors that determine the steady-state level of customers a business will support [2].

Thus, to solve this complicated business problem we decided to make use of neural networks. The fastest implementation from idea to code was through MATLAB [3] and its Neural Network Toolkit. The neural network toolkit provides a pattern matching neural network for which parameters can be specified.

## II. LITERATURE SURVEY

The highlight of the paper is the idea of using the normalized data as an image and using the generated image as input to the neural network. This concept was inspired by the paper on churn data model viewed as periods of time [4].

### A. Dataset Used for the Experiment

Our experiment required data to be processed and used as an input. This was obtained online from bigml [5]. We, thus, obtained a labelled dataset, which enables us to use supervised learning approach for our neural network. Here, our sample dataset appears to be monthly rather than weekly as compared to the dataset used by Wangperpong et al. [4]. Our sample dataset consists of rows - customers and columns as attributes associated with customers. The attributes, which are significant, are not identified in our dataset, but we only consider numeric values while trying to predict the churn rate.

### B. Pattern Matching and Clustering Neural Networks Using Supervised Learning

Furthermore, there has been a lot of work done in pattern matching via supervised learning. Schwenker et al. [6] describe this concept very clearly in their detailed analysis of the topic. This approach to supervised learning is heavily applied to neural networks [7], [8]. Neural networks thus find far and wide reaching applications that can have major impacts on society like the traffic camera system in the reference [9].

## III. METHOD

Raw numeric data is first transformed into the grey scale color domain via min-max normalization. Min-max normalization preserves the integrity of data via scaling. Various methods exist for normalization [10], but the proposed method requires scaling to an interval between [0, 255] hence, we use min-max normalization. Eq. 1 describes min-max normalization, which is a step in the proposed method.

$$X' = a + \frac{(X - X_{\min})(b - a)}{X_{\max} - X_{\min}}. \qquad (1)$$

Here, $X'$ is the normalized pixel. $X$ is the current pixel in consideration. $X_{\max}$ and $X_{\min}$ is the maximum and minimum value of the dataset respectively, a and b are the minimum and maximum values of the color space. Since, the color space is 8 bit we scale the dataset to the interval $[0, 255]$ where a is 0 and b is 255. Hence, we get the following equation:-

$$X' = \frac{(X - X_{\min})255}{X_{\max} - X_{\min}}. \qquad (2)$$

Thus, we get the dataset normalized into the greyscale color space. We, then convert the processed data set into images depending on the time scale. Eq. 2 provides us with a direct formula for converting textual data into a grayscale image.

The time scales are very important to consider along with churn prediction. As mentioned earlier, the dataset is monthly rather than weekly, so a rolling 7-day analysis cannot be made in this case. We can prepare individual images for a customer where a pixel intensity is given by the processed normalized value from the dataset. We generate an image for the entire dataset as evident in Fig. 1. All processing is performed in MATLAB [3]. We also generate a 3D surface plot to describe the entire dataset in three dimensions for greater visualization as shown in Fig. 2.

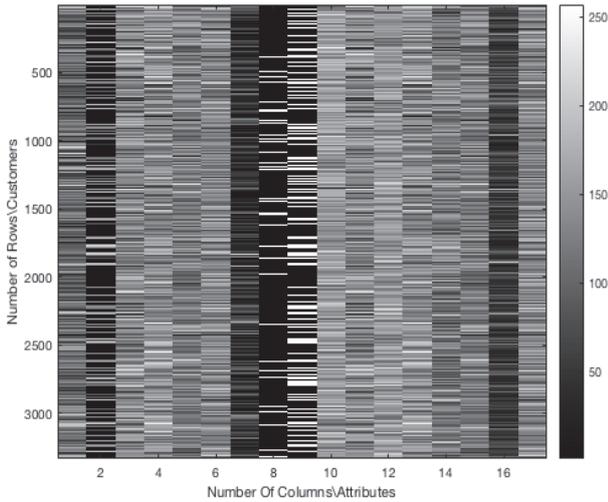

Fig. 1. Customer records and attributes as an image.

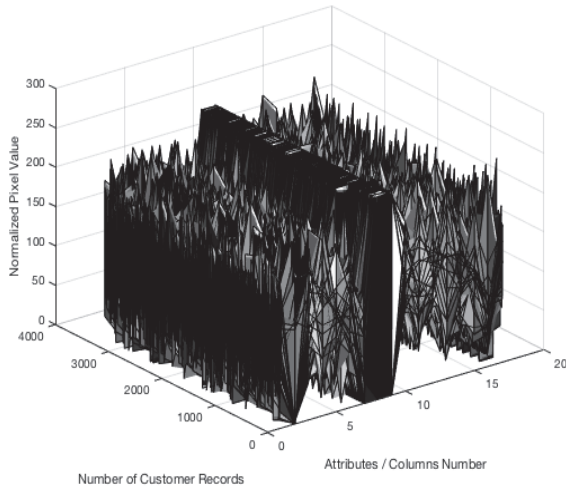

Fig. 2. 3D surface plot of customer records and attributes.

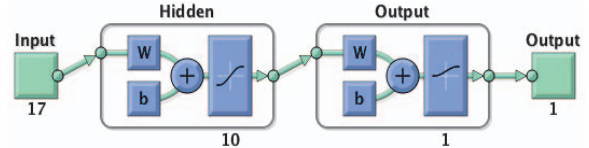

Fig. 3. Structure of neural network.

Fig. 1 describes the processed customer data as an image in grayscale. Here, there is no information loss as long as we have access to the fixed constants $X_{\min}$ and $X_{\max}$ for the dataset under consideration. We can easily calculate the original data by solving the Eq. 2. We use this image for further processing. Fig. 1 represents the image that is generated and used for input to the neural network.

The 3D surface plot shown in Fig. 2 brings to light the pixel intensities across columns for every customer. The sample dataset consists of 3334 customer records with 17 numeric columns. This data excludes the *churn* label column, which describes whether the customer in question has churned. The classification is a binary one where a churn is recorded as 1 and 0 is a retention. We try to predict churn via the neural networks.

The neural network is implemented with the Neural Network Toolkit (NNT) [11] in MATLAB [3]. The NNT contains predefined types of neural networks for clustering, fitting, pattern recognition, and time series. These types make it possible to instantly deploy the neural networks, which otherwise would take considerable time to set up. The NNT handles all initializing of weights and other trivial processes. We use the pattern recognition pre-set for the neural network so that we can detect trends and patterns via neural networks.

As we mentioned earlier, we are using the supervised learning approach since we have a labelled dataset for churn analysis. For creating the neural network, we have to select certain parameters such as < parameter $x$, parameter $y$ >. We have to select the number of hidden layers. This can have an impact on the accuracy of prediction that is obtained. We use 10 hidden layers, which is the default number of layers.

The selected pre-set pattern recognition neural network in the neural network is implemented with the Neural Network Toolkit (NNT) [11] in MATLAB [3]. The NNT contains predefined types of neural networks for clustering, fitting, pattern recognition, and time series. These templates make it possible to instantly deploy neural networks, which otherwise would take considerable time to set up. The NNT handles all initialization of weights and other trivial processes. are handled by the NNT.

NNT has to be trained like any other network for creating a model. We use Scaled conjugate gradient back propagation for supervised training [12]. This is the default algorithm for the pattern recognition neural network in the NNT. The neural network as illustrated in Fig. 3 has 17 inputs, since the input customer attributes are 17. It gives a binary output as evident in Fig. 3.

We observed varying accuracies since the weights are all initialized randomly. Anybody wishing to recreate the experiment may not get exact matching results due to this randomness. But, the accuracy of prediction should be nearby in the neighbourhood of 1–2% or more.

## IV. DISCUSSION OF RESULTS

We obtained an overall accuracy of 94.9% in predicting churn. Furthermore, a comparison of CPU vs. GPU execution is given for the image transformation. All time mentioned is in seconds. The MacBook Air does not have a NVIDIA GPU and MATLAB [3] does not support non – NVIDIA GPUs for graphics acceleration. Hence, the performance numbers for that machine for the GPU domain is not considered. Consider the following figures to explain and further elucidate the results.

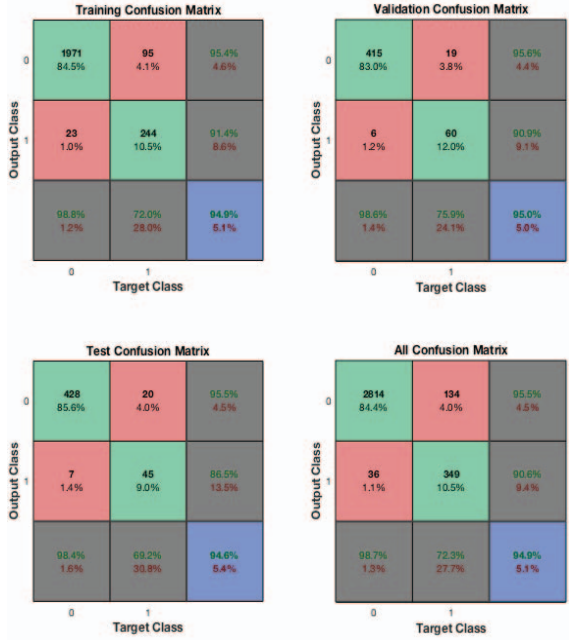

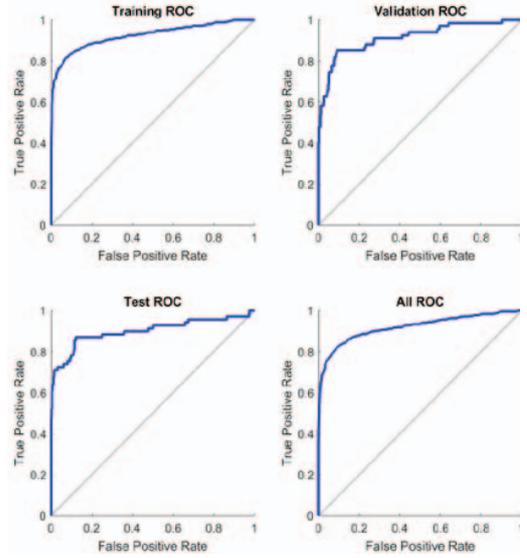

Fig. 6. Receiver operating characteristic (ROC) curves.

Fig. 4. Confusion plot for neural network.

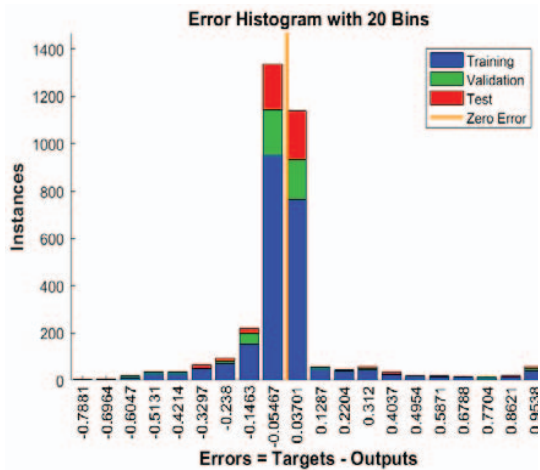

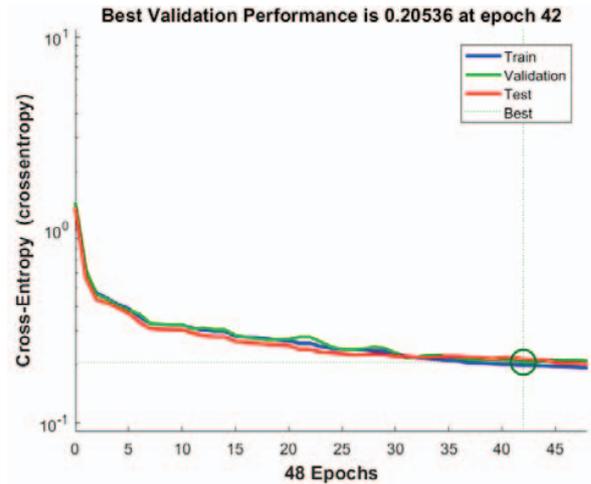

Fig. 5. Error histogram for the neural network.

Fig. 7. Performance plot for the neural network.

The confusion plot in Fig. 4 describes the various results obtained for the different case scenarios including training, validation, and testing. We, thus, observe an overall accuracy of 94.9% in predicting churn. Figs. 5 and 6 provide the error histogram and ROC curves of the Neural Network respectively.

Figs. 7 and 8 describe the performance and training parameters of the neural network. The best validation performance is obtained at epoch 42. The performance can be contrasted with machines having two different Operating Systems and Hardware. The Table I presents numbers that explain very clearly the performance difference between systems that can be appreciably measured.

## V. EXPERIMENTAL COMPUTER CONFIGURATION

Compute Unified Device Architecture (CUDA) support in graphics cards or processors, popularly known as Graphics Processing Unit (GPUs). CUDA accelerates computational tasks and require less time to execute [12]. The above discussed image transformation is executed on two systems of different configuration.

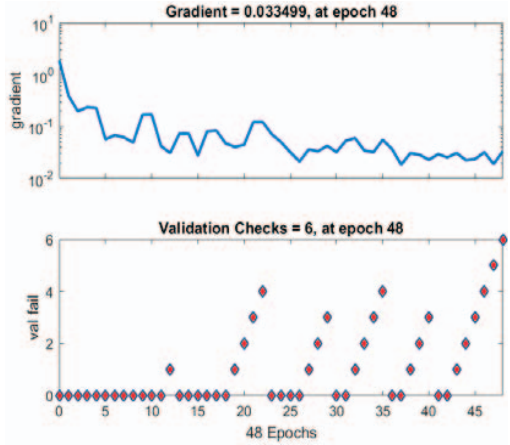

Fig. 8. Training state for the neural network.

TABLE I
EXECUTION TIME ON EXPERMENTAL SYSTEMS.

| Computer description | CPU execution time (seconds) | GPU execution time (seconds) |
|---|---|---|
| High End Custom PC | 0.001553 | 0.001051 |
| MacBook Air Early 2015 | 0.004755 | N/A[a] |

a. Does not have CUDA GPU.

TABLE II
SYSTEM CONFIGURATION FOR EXPERMENTAL SYSTEMS.

| Processor | Physical RAM | CUDA GPU name | MATLAB GPU support | Operating system and build |
|---|---|---|---|---|
| Intel Core i7 6700k @ 4 GHz | 2 × 8 GB Kingston DDR4 Non ECC @ 2133 MHz | NVIDIA Titan X Maxwell Architecture 12 GB GDDR5 | Yes | Windows 10 64 bit Build 1607, Custom PC Build |
| Intel Core i5 @ 1.6 GHz | 4 GB DDR3 Non ECC @ 1600 MHz | N/A[a] | No | MacOS Sierra 10.12.2 Build 16C67 on MacBook Air Early 2015, 13 inch |

a. Does not have CUDA GPU.

As evidenced in Tables I and II different computer configurations lead to varying performance numbers and lead to interesting comparisons between possible hardware and software choices. A mobile machine like a MacBook Air that although manages to perform the necessary task, it is clearly not made for heavy and intensive scientific work. The numbers in the Table II clearly highlight this fact.

## VI. CONCLUSIONS

We obtained a method for processing numeric data as image input to a pattern matching the neural network. This provides a new strategy of analysing entire numeric datasets as images. This opens up a whole new perspective at data which can be visually conceptualised as an image viewable on a display or screen. Visual strategies to data become a possibility if this method is considered. Further the colour domain may be expanded to color spaces other than grayscale. after scaling to the required interval [15].

## VII. FURTHER POSSIBLE WORK

MATLAB was particularly helpful for implementing this experiment. Hence, anyone attempting to replicate the same experiment would be advised to follow a similar approach via MATLAB or any other similar software or framework. To be clear, the actual neural network was not run on the GPU at the moment. A GPU implementation makes the neural networks run faster as compared to CPU implementations, as evidenced in many earlier implementations. Hence, it would be interesting to see how results pan out on the GPU when the neural network is implemented. A larger dataset would have been helpful to evaluate this experiment on a larger scale.

**Acknowledgements:** I would thank Prof. Agrawal for his guidance in developing the correct thought process as my project advisor. I would also thank my friends and team mates Santosh Shettar and Amish Thakkar for encouraging me to develop this idea. Finally, last but the most, my parents for sponsoring my PC and enabling me to complete this experiment and for their unwavering support.